\pdfoutput=1 
\documentclass{bmvc2k}
\pdfoutput=1 
\usepackage{microtype}
\usepackage{graphicx}
\usepackage{booktabs} % for professional tables
\usepackage{amsmath}
\usepackage{amssymb}
\usepackage{multirow}
\usepackage{caption}
\usepackage{algpseudocode}
\usepackage[ruled,vlined]{algorithm2e}
\usepackage{bbold}
\usepackage{mathtools}
\usepackage{wrapfig}
\usepackage{lscape}
\usepackage{rotating}
\usepackage{epstopdf}
\usepackage{amsmath}
\usepackage{physics}

\algnewcommand{\algorithmicforeach}{\textbf{for each}}
\algdef{SE}[FOR]{ForEach}{EndForEach}[1]
  {\algorithmicforeach\ #1\ \algorithmicdo}% \ForEach{#1}
  {\algorithmicend\ \algorithmicforeach}% \EndForEach
\title{Frequency-Based Vulnerability Analysis of Deep Learning Models against Image Corruptions}

\addauthor{Harshitha Machiraju$^\star$}{harshitha.machiraju@epfl.ch}{1,2}
\addauthor{Michael H. Herzog}{michael.herzog@epfl.ch}{1}
\addauthor{Pascal Frossard}{pascal.frossard@epfl.ch}{2}

\addinstitution{
 Laboratory of Psychophysics (LPSY)\\
 Ecole Polytechnique Fédérale de Lausanne (EPFL), \\
 Switzerland
}
\addinstitution{
 Signal Processing Laboratory 4 (LTS4)\\
 Ecole Polytechnique Fédérale de Lausanne (EPFL), \\
 Switzerland
}

\runninghead{Machiraju et al}{MUFIA}

\DeclareMathOperator*{\argmax}{arg\,max}
\DeclareMathOperator*{\argmin}{arg\,min}

\begin{document}

\maketitle

\begin{abstract}
Deep learning models often face challenges when handling real-world image corruptions. In response, researchers have developed image corruption datasets to evaluate the performance of deep neural networks in handling such corruptions. However, these datasets have a significant limitation: they do not account for all corruptions encountered in real-life scenarios. To address this gap, we present MUFIA (Multiplicative Filter Attack), an algorithm designed to identify the specific types of corruptions that can cause models to fail.
Our algorithm identifies the combination of image frequency components that render a model susceptible to misclassification while preserving the semantic similarity to the original image. 
We find that even state-of-the-art models trained to be robust against known common corruptions struggle against the low visibility-based corruptions crafted by MUFIA. This highlights the need for more comprehensive approaches to enhance model robustness against a wider range of real-world image corruptions. \footnote{Code: \href{https://github.com/code-Assasin/MUFIACode}{https://github.com/code-Assasin/MUFIACode}}
\end{abstract}

%-------------------------------------------------------------------------
\section{Introduction}
\label{sec:intro}
\vspace{-2mm}
Deep learning models have achieved remarkable performance in various tasks, such as image classification \cite{imagenet}, object detection \cite{yolo_new}, and segmentation \cite{dino}. However, their efficacy often does not translate to real-world scenarios, where images are affected by various types of corruptions, including weather \cite{bethge_av,mine_av} and style \cite{shape_texture} changes. Hendrycks et al. \cite{cc} built the common corruptions dataset \cite{cc} to benchmark the ability of deep neural networks to handle such real-world image changes.
They introduced image corruptions based on 4 categories of noise, blur, weather, and digital-based image changes. They add these corruptions to the existing CIFAR10/100 \cite{cifar10/100} and ImageNet \cite{imagenet} datasets. They also showed that existing models trained on the clean version of these datasets fail to generalize to their corrupted counterparts. 

To improve the performance of standard models on this common corruption dataset, several methods have been introduced \cite{Prime,Augmix,card,imagenet_sota}. However, these methods are designed by considering only the type of corruptions present in the common corruption dataset \cite{cc}. Mintun et al. \cite{cc_better} created a dataset with corruptions that were different that those in the original common corruption dataset \cite{cc}. They found that many of the models that achieved state-of-the-art performance on the original common corruptions dataset failed to generalize to these different corruptions. 
This showed a fundamental limitation of benchmarking on these corruption datasets: newer corruptions that share no similarity to the existing ones will always be found in real-life scenarios. This makes it very hard to evaluate the performance of different methods in widespread realistic scenarios. For the same reasons, it is impractical to keep building new image corruption datasets to identify new vulnerabilities and achieve robustness in the corresponding settings. \\
One possible approach to simplify the analysis of model robustness against different image corruptions is to characterize these corruptions based on their frequency content. By understanding the frequency biases of models, we can estimate their performance on unseen corruptions with different frequencies. Many works \cite{bethge_freq,fourier_yin,guille,mine_bio} have conducted frequency-based analyses by creating filtered datasets using bandpass filters and testing different deep learning models on these filtered samples. 
However, this approach is computationally expensive as it involves creating datasets for each frequency bandwidth. Moreover, it fails to consider the combined effect of different frequency components in image corruptions. For example, images can be corrupted by a combination of mid-frequency blur and low-frequency brightness changes. Creating new datasets with different combinations of frequencies for every possible corruption is not suitable. Hence, it would be more feasible to develop an algorithm that can identify the combination of frequencies that the model is most vulnerable to. This would eliminate the need for creating new corruption datasets and permit to identify the types of real-world corruptions associated with specific frequency combinations that are problematic for image models.\\
In this work, we propose a new algorithm called MUFIA (\textbf{Mu}ltiplicative \textbf{Fi}lter \textbf{A}ttack) that creates an adversarial and differentiable frequency filter bank in the Discrete Cosine Transform (DCT) domain \cite{dct_orig}. For a given image and model, we find the combination of frequency components that produce an adversarial image. We do this while maintaining high semantic similarity to the original image. This combination of frequency components represents an adversarial filter bank and specially characterizes the combination of frequencies that the model is most vulnerable to. By analyzing these adversarial images, we can identify the types of corruptions where the model is likely to fail and the corresponding real-life scenarios.
Through extensive experiments, our work reveals that most data augmentation-based methods \cite{Prime,Augmix} used on the common corruption dataset \cite{cc} are vulnerable to a combination of blur and brightness changes. This suggests that these augmentation techniques are incomplete and those models remain susceptible to such corruptions. Furthermore, we find that even state-of-the-art models trained to be robust against known corruptions like fog struggle against the low visibility-based corruptions identified by our attack algorithm. Our work thus clearly highlights the need for more comprehensive approaches to enhance model robustness against a wider range of real-world image corruptions.
\vspace*{-6mm}
\section{Related Work}
\vspace{-3mm}
\label{sec:related_work}
In this Section, we introduce the common corruption datasets in more detail and then follow up on works that create adversarial images.
\label{sec:ref}
\vspace{-4mm}
\subsection{Common Corruptions Dataset}
\vspace{-2mm}
The Common Corruptions (CC) dataset, introduced by Hendrycks et al. \cite{cc}, is a benchmark dataset designed to evaluate the ability of deep neural networks to handle real-world image corruptions. The dataset includes 15 different types of image corruptions categorized into four main groups: noise, blur, weather, and digital-based changes. Examples of corruptions include Gaussian noise, motion blur, fog, and compression artifacts. These corruptions are randomly added to existing datasets such as CIFAR10/100 \cite{cifar10/100}, and ImageNet \cite{imagenet}, thereby creating corrupted versions of these datasets called CIFAR10-10/100-C and ImageNet-C. The main concept behind the CC dataset is to prescribe a benchmark framework when models are not directly trained on the specific corruptions present in CC. Then, many works have come up with alternate training methods to generalize well to this dataset. Some of the popular ones include:
\vspace{-2mm}
\begin{itemize}
\itemsep-0.5em 
    \item \textbf{Card} \cite{card}: This method is currently the state of the art accuracy on CIFAR10-C and CIFAR100-C. It utilizes an ensemble of networks pruned to $95\%$ sparsity while maintaining high classification accuracy. The authors hypothesize that compressing the model prevents overfitting and improves generalization properties.
    \item \textbf{DeiT-B} \cite{imagenet_sota}: Data-efficient Image Transformers (DeiT) represent a  class of vision transformer-based models that are trained in a self-supervised manner. The DeiT-B model achieves the current state-of-the-art accuracy on ImageNet-C.
    \item \textbf{Prime} \cite{Prime} and \textbf{Augmix} \cite{Augmix}: These are popular defense methods based on data augmentation. Prime \cite{Prime} applies random spatial, spectral, and color-based transformations, while Augmix utilizes random translation, rotation, and posterizing \cite{auto_augment} transformations to augment the data.\footnote{All models were obtained from the RobustBench leaderboard \cite{robustbench}.}
\end{itemize}
\vspace{-6mm}
\subsection{Adversarial Filter Bank}
\vspace{-2mm}

Adversarial attacks typically produce highly imperceptible additive perturbations bounded in $L_p$ norm \cite{carlini_attack,madry_adv_train}. However, more recently, there have been several adversarial attacks that produce semantically similar images based on color that are not bounded in $L_p$ norm \cite{ace_color,texture_attack}. 
In addition, these attacks \cite{ace_color,texture_attack} fall short in capturing the magnitude of changes seen in real-world corruptions \cite{guille_survey, maksym_adv_cc}, thus limiting their implications to real-world scenarios.

However, none of these works utilize the frequency domain based changes like blurs, brightness changes etc. 

 Our algorithm, MUFIA (Multiplicative Filter Attack), addresses the limitations of existing frequency-based analyses for common corruption datasets \cite{guille, fourier_yin, mine_bio}. MUFIA leverages an adversarial frequency filter bank for misclassification, akin to $L_p$-norm based adversarial attacks \cite{carlini_attack, madry_adv_train}.
 By exploring the frequency domain and analyzing specific frequency combinations, Our method uncovers realistic corruptions, revealing new vulnerabilities of deep learning models.
To the best of our knowledge, our work stands as the first in the image domain to produce a multiplicative adversarial filter bank, as detailed in the following section.

\vspace{-4mm}
\section{Methodology}
\vspace{-3mm}
\label{sec:methodology}
In this Section, we introduce the algorithm of MUFIA and subsequently delve into the loss functions used in creating our adversarial filter bank.
\vspace{-3mm}
% \subsection{Filter Bank}
\subsection{MUFIA : Multiplicative Filter Attack}
\vspace{-2mm}
Our attack algorithm, described in Algo.\ref{algo:original}, begins by converting RGB images to the YCbCr domain, separating the luma (Y) and chroma components (CbCr). We focus exclusively on the luma components while keeping the chroma components unchanged. This approach ensures that the resulting adversarial images exhibit more realistic changes, while avoiding unrealistic color alterations such as a bright neon sky. 
\footnote{In tasks like semantic segmentation, edge detection, depth estimation, optical flow, etc., where Chroma information is not crucial, our algorithm can be adapted to incorporate filter banks on these channels as well. However, for image classification, introducing unrealistic color changes holds little practical significance in evaluating the model. Moreover, focusing on variations in the Luma components alone provides valuable insights (as shown in Sec. \ref{sec:experiments}) for our specific research.}

\begin{algorithm}[H]
\SetAlgoLined

\begin{flushleft}
        \textbf{Input:} $X \xleftarrow{}$ Dataset of images  \\
        $C \xleftarrow{}$ One-hot encoded dataset of labels  \\
        % $B\xleftarrow{}$  Batch Size \\
        $N\xleftarrow{}$  Patch size/Block size for DCT algorithm. \\
        % $Y \xleftarrow{}$ Batch of true labels \\
        % $Q \xleftarrow{}$Filter Bank, initialized with $\mathbf{1}^{D \times N \times N}$\\
        % $B\xleftarrow{}$ Batch size \\ 
        $N_{iters}\xleftarrow{}$ Number of iterations of the attack \\
        $D \xleftarrow{} $ Type-II DCT operator for block size $N$ \\
        $\hat{X} \in \{\}  \xleftarrow{}$ Empty set of adversarial images\\
        $B\xleftarrow{}$  Number of blocks generated by dividing an image, $x \in X$, into blocks of size $N \times N$  \\

        \textbf{Output:}  $\hat{X} \xleftarrow{}$ Adversarial images \\
        % ${Q}\xleftarrow{}$ Adversarial filter banks.
\end{flushleft}

 \For{$ x \in X$ and $ c \in C$}
 {
    Initialize Filter bank for $x$ $\xrightarrow{} Q \in \mathbf{1}^{N \times N} $ \\
    Initialize:  $\hat{x} = x$ \\
    Convert $x$ from RGB to YCbCr $\xrightarrow{} \mathbf{x}$ \\
    Separate $\mathbf{x}$ into channels $\xrightarrow{} \mathbf{y,cb,cr}$ \\
    Compute block wise DCT of Y channel:    $ D(\mathbf{y}) \rightarrow  \{ y_1 \dots y_B \}$ \\
    \For{$i \in N_{iters}$}{
    
        Update $Q$ : $\argmin_{Q} ( \mathcal{L}_{adv} (\hat{x}, c) + \lambda \mathcal{L}_{sim}(\hat{x},x) ) $ \Comment{Sec.\ref{sec:attack}} \\
         \For{$b \in B$}{
            $\hat{y}_b = Q \odot y_b$ 
         }
        Compute inverse DCT and recombine blocks : $D^T(\{ \hat{y}_1 \dots \hat{y}_B \}) \rightarrow \hat{\mathbf{y}}$  \\
        Recombine channels: $\{\hat{\mathbf{y}}, \mathbf{cb}, \mathbf{cr}\} \rightarrow \hat{\mathbf{x}} $ \\
        Convert $\hat{\mathbf{x}}$ from YCbCr to RGB $\xrightarrow{} \hat{x} $ 
        % Clamp $\hat{x}$ to [0,1] \\
        % Calculate $\mathcal{L}_{adv}$ =  $\max(\mathcal{S}_{cos}(f(\hat{x}), y) + \kappa, 0)$ \\
        % Calculate $\mathcal{L}_{iqa}$ = $1 - \cos(\hat{y}_{orig}, \hat{y}_{new})$ \\
        
    }
    Update : $\hat{X} \gets \hat{X} \cup \{\hat{x}\}$
 }
\caption{\textbf{MU}ltiplicative \textbf{FI}lter \textbf{A}ttack (MUFIA)}
\label{algo:original}
\end{algorithm}
% \vspace{-3mm}

We initially divide the Y channel of the image into smaller blocks of size $N \times N$ and apply the Discrete Cosine Transform (DCT) operation. By employing a filter bank in the DCT domain, we can amplify or attenuate different frequencies in the image. This is accomplished through element-wise multiplication of the frequency coefficients with a matrix $Q\in \mathbb{R}^{N \times N}$, representing the desired alteration degree for each frequency band. We update $Q$ based on the loss terms defined in Sec.\ref{sec:attack}, ensuring it induces misclassification in the model for the given image $x$, while producing a semantically similar adversarial image $\hat{x}$. The filter bank ($Q$) is then uniformly applied to all DCT blocks, ensuring consistent behavior across the entire image.\footnote{While an alternative approach would involve designing separate filter banks for each block, such an approach significantly increases computational complexity. Additionally, our chosen method guarantees homogeneous content across all blocks and, consequently, in the final image.} When $Q \in \mathbf{1}^{N \times N}$, the original image is obtained. Scaling $Q$ may introduce visible blurring, brightening, or sharpening effects in the image.\footnote{Throughout the paper, we refer to $Q$ as a filter bank, although technically it works in conjunction with the DCT transform as a filter bank. Note that a similar principle is employed in JPEG-type compression schemes, where the matrix $Q$ is represented by the quantization matrix \cite{jpeg_orig}.} Finally in our algorithm, after applying the filter bank to the Y channel DCT blocks, we return to the image domain by performing the inverse DCT, followed by conversion back to the RGB color space. This algorithm is repeated for multiple iterations to find an adversarial filter bank $Q$.
\vspace{-2mm}

\subsection{Losses}
\label{sec:attack}
\vspace{-2mm}
We now examine the loss terms necessary to establish the adversarial nature of our filter bank $Q$. To achieve this, we employ an adversarial loss function that optimizes $Q$ to ensure misclassification. Thus, our framework is structured as follows: given a classifier network $f$ and an input image $x \in [0, 1]^{H \times W \times Ch}$, where $H$, $W$, and $Ch$ represent the height, width, and number of channels in the image respectively, and $c$ represents the true label of the image $x$ encoded in a one-hot format, we aim to generate an adversarial image $\hat{x}$ as outlined in Algo.\ref{algo:original} with the help of a filter bank, $Q$, by minimizing an adversarial loss term as:
\vspace{-1mm}
\begin{gather}
    \min_{Q} \mathcal{L}_{adv} =  \min_{Q} ({  \max(\mathcal{S}_{cos}(f(\hat{x}), c) + \kappa, 0) }) 
\label{eq:loss_adv}
\end{gather}
% \vspace{-2mm}
Here, $\mathcal{S}_{cos}(,)$ represents the cosine similarity between the two vectors, and $\kappa$ is the minimum margin. To analyze our loss term, we first look at the inner part
${\mathcal{S}_{cos}(f(\hat{x}), c)}$ that contributes to minimize the cosine similarity between the prediction of the network (logits) and the true class label. Our loss term thus enforces the logits of the model to be maximally misaligned with true class direction hence causing misclassification. This is loosely inspired by the adversarial loss used for $L_p$ norm adversarial attacks in the audio domain \cite{audio_loss_cite,voicebox}. \\
The complete loss term in Eq.\ref{eq:loss_adv} consists of an additional hinge-loss-like constraint ($max(;0)$). The hinge-loss term is inspired by the Carlini-Wagner attack \cite{carlini_attack} and ensures a higher penalty when the model has confidence in the true class. This loss term on its own does not guarantee misclassification. To ensure that the cosine similarity ($\mathcal{S}_{cos}(,)$) is at least reduced by a certain margin, we introduce the parameter $\kappa$. With higher $\kappa$ values, we enforce the attack to have larger misclassification rates. Since, the above adversarial loss $\mathcal{L}_{adv}$ purely focuses on ensuring misclassification, thus it is possible that our attack makes the image completely unrecognizable. To prevent this, we add a regularization term to ensure that the DCT coefficients of the original image $x$, and those of the adversarial image $\hat{x}$ are still similar. We define this similarity term as follows:
\vspace{-2mm}
\begin{gather}
   \mathcal{L}_{sim} = 1 - \mathcal{S}_{cos}(D(\hat{x}), D(x)) 
   \vspace{-5mm}
\label{eq:sim}
\end{gather}
Combining Eq.\ref{eq:loss_adv} and \ref{eq:sim}, we write our total loss term as follows:
\vspace{-1mm}
\begin{gather*}
   \mathcal{L}_{total} =  \mathcal{L}_{adv} + \lambda \mathcal{L}_{sim} 
   \vspace{-2mm}
\end{gather*}
By varying the hyperparameter $\lambda$, we can regulate the trade-off between producing highly semantically similar images and causing larger misclassification rates. 

% \vspace{-4mm}
\section{Experiments}
% \vspace{-2mm}
\label{sec:experiments}
\subsection{Settings}
\vspace{-1mm}
Through our attack, we want to characterize a new class of corruptions that various models are vulnerable to. For this purpose, we conduct evaluations on three datasets: CIFAR10, CIFAR100 \cite{cifar10/100}, and ImageNet \cite{imagenet}. 
For the models, we include Card \cite{card}, Prime \cite{Prime}, Augmix \cite{Augmix}, and DeiT \cite{imagenet_sota}, the details of which are mentioned in Sec.\ref{sec:related_work}. For a baseline comparison, we also include standard models which were only trained on the clean dataset. We used the Adam optimizer for all our experiments with a learning rate of 0.1 and a batch size of 32. Through experimentation, we found that settings $\kappa = 0.99$, $\lambda = 20$, and $N_{iters} = 100$ iterations for our attack yielded good results across all our experiments. 
Further, we use a filter bank of size 32 and 56 for CIFAR-10/100 and ImageNet respectively. We analyze these hyperparameters in our Supplementary work (Sec. \ref{sec:Ablation}). 
% \vspace{-4mm}

\begin{table}[h!]
\centering
\resizebox{0.9\textwidth}{!} {
\begin{tabular}{@{}c|c|c|c|c|c|c@{}}
\toprule
\multirow{2}{*}{Dataset} & \multirow{2}{*}{Model} & \multirow{2}{*}{Architecture} & \multicolumn{3}{c|}{Accuracy ($\%$)} & \multirow{2}{*}{LPIPS ($\downarrow$)} \\ \cline{4-6}
                         &                        &                               & Clean     & CC    & MUFIA       \\ \midrule
    \multirow{4}{*}{CIFAR10} & \textsc{Standard} \cite{resnet} & ResNet-50 & 95.25 & 73.46 & 0 & 0.076\\
                        & \textsc{Prime} \cite{Prime} & ResNet-18 & 93.06 & 89.05 & 2.27 & 0.163\\
                        &  \textsc{Augmix} \cite{Augmix} & ResNeXt29\_32x4d & 95.83 & 89.09 & 0 & 0.119 \\
                        &\textsc{CadD} \cite{card} & WideResNet-18-2 & \textbf{96.56} & \textbf{92.78} & \textbf{38.71} & 0.181 \\ \midrule
\multirow{4}{*}{CIFAR100}    & \textsc{Standard} \cite{resnet} & ResNet-56 & 72.63 & 43.93 & 0 & 0.09\\
                        &\textsc{Prime} \cite{Prime} & ResNet-18 & 77.60 & 68.28 & 0.76 &  0.132\\
                        &\textsc{Augmix} \cite{Augmix} & ResNeXt29\_32x4d & 78.90 & 65.14 & 0.05 & 0.1536 \\
                        &\textsc{Card} \cite{card} & WideResNet-18-2 & \textbf{79.93} & \textbf{71.08} & \textbf{13.31} & 0.148 \\  \midrule   

\multirow{4}{*}{ImageNet} & \textsc{Standard} \cite{resnet} & ResNet-50 & 76.72 & 39.48 & 0.712 & 0.196\\
                        & \textsc{Prime} \cite{Prime} & ResNet-50 & 75.3 & 56.4 & 0.45 &  0.224\\
                        & \textsc{Augmix} \cite{Augmix} & ResNet-50 & 77.34 & 49.33 & 0.438 & 0.243 \\
                        & \textsc{Deit-B} \cite{imagenet_sota} & DeiT Base & \textbf{81.38} & \textbf{67.55} &\textbf{ 6.748} & 0.221\\  \bottomrule
\end{tabular}
}
\caption{Accuracy ($\%$) on the clean uncorrupted dataset (clean), common corruptions dataset (CC), and accuracy after our attack (MUFIA) for standard and robust models on different datasets. Our attack algorithm, MUFIA, exposes previously unseen corruptions that greatly impact the accuracy of almost all the models. Notably, while these models perform well on the common corruption dataset (CC accuracy column), they struggle when confronted with new corruptions introduced by MUFIA. Further, our attack achieves this while generating adversarial images that maintain a high degree of semantic similarity, as indicated by the LPIPS values \cite{lpips}. \vspace{-4mm}}
% \caption{Results on different models trained on CIFAR-10 and its variants.. We label the model with standard training as $Std$ and the ones adversarially trained with $\epsilon_{L_{2}} =0.5$ and $\epsilon_{L_{\infty}}= \frac{8}{255}$ PGD attacks as $L_{2}$ and $L_{\infty}$ models respectively. }
\label{tab:res_all}
\end{table}

\subsection{Results}
We use the MUFIA algorithm (described in Sec. \ref{sec:methodology}) to assess the accuracy of the mentioned models on the novel corruptions introduced by our attack. The results are presented in the MUFIA column of Table \ref{tab:res_all}. As a baseline, we also included the accuracy of the models on the common corruption datasets (denoted as CC), which represents the average accuracy across all corruptions mentioned in Sec. \ref{sec:related_work}. Additionally, we provided the accuracy of the models on the clean, uncorrupted datasets in the Clean column. To evaluate the average image similarity of the images produced by MUFIA, we employed LPIPS \cite{lpips}, which measures the perceptual distance between images. Lower LPIPS values indicate higher semantic similarity. Notably, Kireev et al. \cite{maksym_adv_cc} reported that the maximum LPIPS distance observed on the common corruption datasets is approximately 0.45, while the average is around 0.13. In Table \ref{tab:res_all}, our algorithm consistently achieves LPIPS values within a comparable range, indicating that our approach exhibits a high degree of semantic similarity in accordance with the widely recognized Common Corruption dataset \cite{cc}.

% \clearpage

\begin{figure}[t!]
    \centering
        \includegraphics[width=\textwidth]{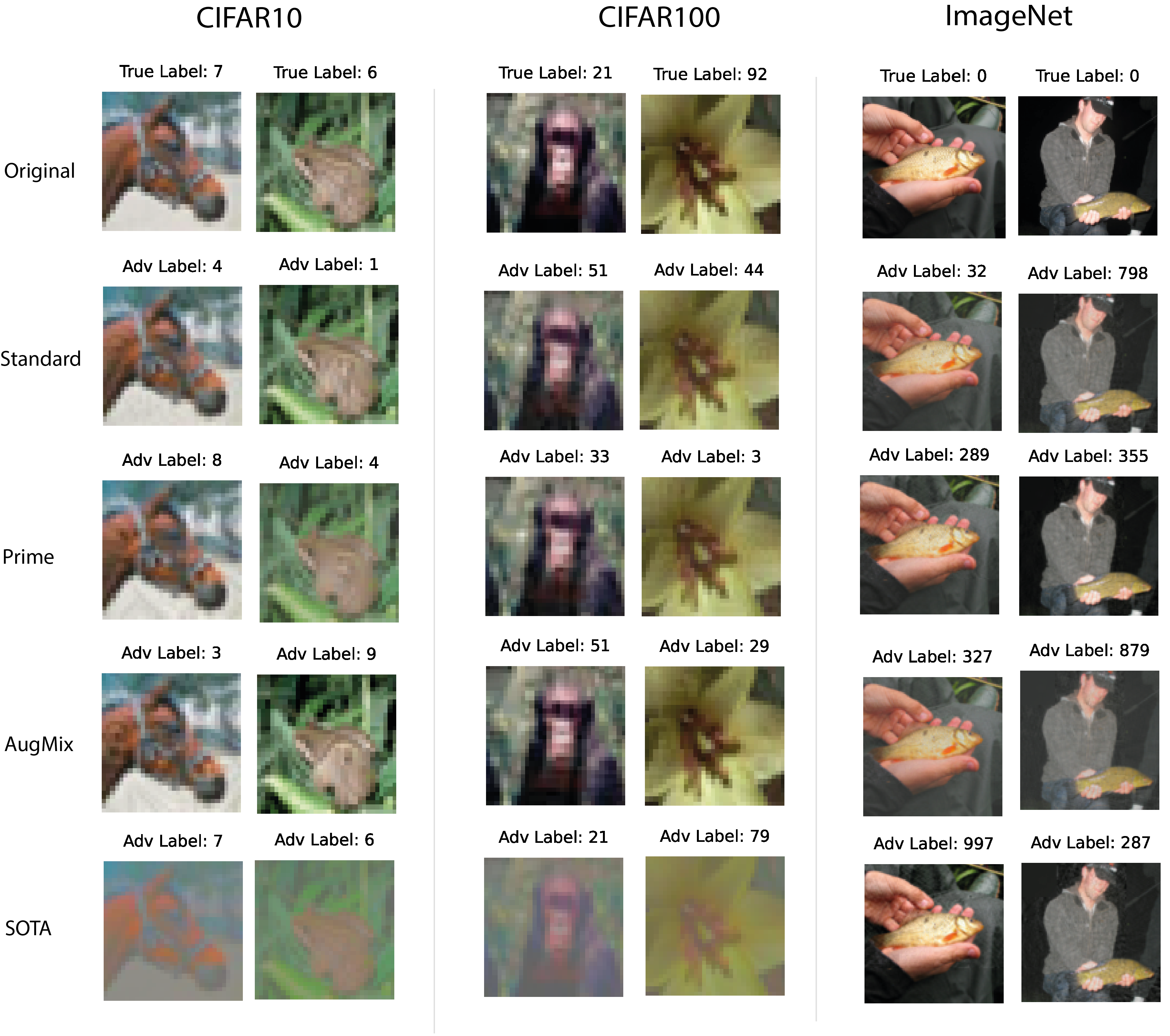}
    \caption{Images generated by our attack algorithm, MUFIA, for different datasets and models. Each column represents a dataset, and the first row shows the original, unchanged images with their true labels. The following rows show the images generated by MUFIA for different models, with the predicted label indicated at the top of each adversarial image. Note, for the CIFAR-10/100 columns, the SOTA model corresponds to Card \cite{card}, and for the ImageNet column, it corresponds to the DeiT-B model \cite{imagenet_sota}. To observe the subtle differences in the images themselves, we encourage the readers to zoom in. We can observe the differences between the original and adversarial images, such as lower visibility for the SOTA model on CIFAR10/100 (Card \cite{card}) and a soft blurring and decreased brightness effect for all models except for the SOTA model on ImageNet (Deit-B \cite{imagenet_sota}). 
    Furthermore, the fisherman image (last column on the right) showcases a slight increase in brightness and blurring across all models, with an additional shadow figure visible due to the heightened brightness. }
    % \vspace{-30pt}
    \label{fig:samples_filters}
\end{figure}

Moreover, in Fig.\ref{fig:samples_filters}, we showcase examples of adversarial images generated by our attack. Our attack evidently induces subtle blurring or brightness changes while maintaining high semantic similarity to the original samples. 
\\ \\
Table \ref{tab:res_all} reveals the accuracy of different models when subject to our attack. Notably, our attack successfully compromises the Standard, Prime \cite{Prime}, and Augmix \cite{Augmix} models while significantly reducing the accuracy of state-of-the-art models, Card \cite{card}, and DeiT-B \cite{imagenet_sota}. It is noteworthy that both Prime \cite{Prime} and Augmix \cite{Augmix} are designed to be robust to changes like blur and brightness found in the common corruption dataset (see CC column in Table. \ref{tab:res_all}). However, they are easily deceived by changes introduced by our attack (as observed in Fig.\ref{fig:samples_filters}). This shows that these data augmentation methods may not generalize very well to new corruptions present in realistic scenarios.
However, as seen in Table \ref{tab:res_all}, the state-of-the-art model on CIFAR-10/100 (Card \cite{card}) still exhibits some robustness against our attack. To gain further insights into this performance, we analyze the filter banks produced by our attack for all these models.

\begin{figure}[t!]
    \centering
    % \captionsetup{format = hang}
    \includegraphics[width=\textwidth]{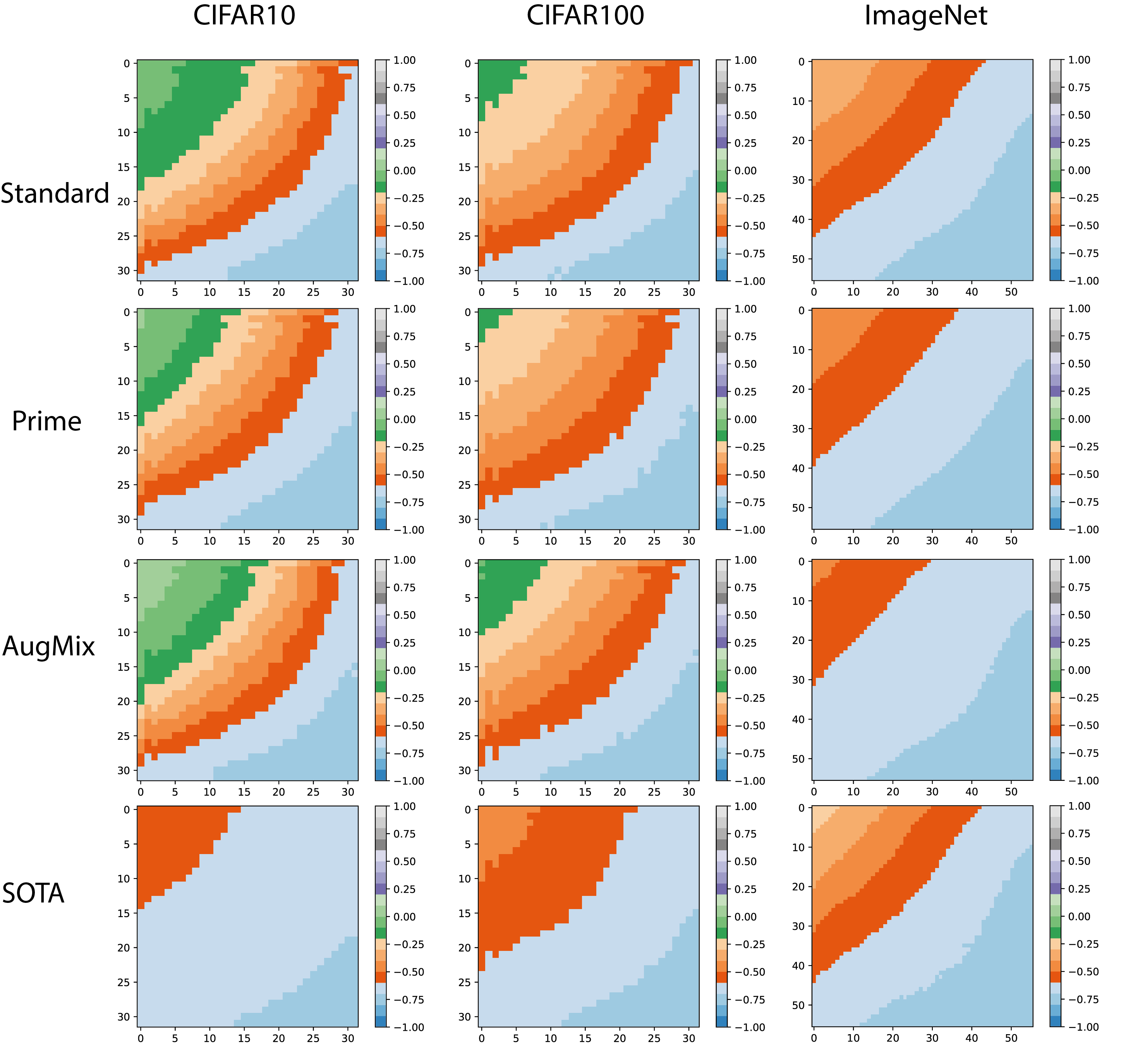}
    \caption{The visualization shown in Fig.\ref{fig:median_filter} depicts the heatmap generated by our algorithm MUFIA, which represents the median of all filter banks. These filter banks are calculated for a given dataset and model. In the figure, each column represents a dataset, and each row corresponds to a model evaluated by our attack algorithm, MUFIA. Note, for the CIFAR-10/100 columns, the SOTA model corresponds to Card \cite{card}, and for the ImageNet column, it corresponds to the DeiT-B model \cite{imagenet_sota}.
    To facilitate visual comparison, we apply a transformation using the $tanh(z-1)$ function to map the filter banks within the range of $(-1,1)$. In this transformed representation, a value of $0$ (green in the heatmap) indicates an unchanged frequency component $(z=1)$, positive values indicate amplification $(z>1)$ (purple, grey in the heatmap), and negative values indicate attenuation $(z<1)$ (orange/red, blues in the heatmap). 
    Notice that for all the models, the high frequencies are always altered. However, for CIFAR10/100, the lower frequencies are significantly more preserved compared to ImageNet. Hence introducing different changes in the images as seen in Fig.\ref{fig:samples_filters}.}
    \label{fig:median_filter}
    % \vspace{-30pt}
\end{figure}

% \red{ADD THAT CC IS AVERAGE FOR BLUR< BRIGHTNESS ETC SO EXPECTED IS QUITE HIGH BUT ON OUR ATTACK ITS LOW The evaluation metrics used on this benchmark dataset is the average accuracy across all corruptions of the dataset at various severity levels. Ther ei}

% \vspace{-4mm}
 \subsection{Filter Bank}
% \vspace{-7mm}
Through the analysis of filter banks generated for a model, we can further understand its frequency bias and the corruptions it is vulnerable to.
Intuitively, to maintain high image similarity while trying to cause misclassification, one might think of altering the higher frequency components. Diminishing these high-frequency components can lead to a loss of information such as edges and texture, hence causing the model to misclassify (similar to the effect of blurs in the common corruptions dataset). Conversely, if the model remains robust, it indicates that altering the lower frequency components is necessary for misclassification. This can result in overall changes in brightness. Reducing the brightness can often decrease the visibility of the object and hence successfully deceive the model (similar effect as for fog, brightness, and contrast in the common corruptions dataset). 
Therefore, for a given model, by looking at the median of the filter banks generated for all the samples in the dataset, we can determine the type of frequency bias that it contains. If our attack can successfully cause misclassification of the model, while its median filter bank shows that most of the lower frequencies remain relatively unchanged, it suggests that the model is biased toward high frequencies. Conversely, if the model's median filter bank also has changes in many low frequencies, it implies that it does not have a specific bias to any frequency but rather needs a combination of frequencies to cause misclassification. Despite these changes, when the model is robust to our attack, it demonstrates excellent object recognition capabilities, which would translate well to real-world corruptions. In order to study this, we visualize the median of the filter bank for each dataset and model in Fig.\ref{fig:median_filter}. \\ 
\textbf{CIFAR-10/100 :} In Fig.\ref{fig:median_filter}, we observe the characteristics of filter banks for CIFAR-10/100. Except for the Card model \cite{card}, other models' median filter banks show lesser modification (\textcolor{orange}{orange}) (sometimes even unchanged (\textcolor{green!70!black}{green})) of low-frequency components compared to high-frequency counterparts (\textcolor{cyan!40}{blue}). This indicates a bias towards high frequencies in these models. Similar observations were made by Yin et al. \cite{fourier_yin} and Ortiz et al. \cite{guille} for the baseline standard models.
In contrast, the median filter bank of the state-of-the-art Card model \cite{card} does not preserve low frequencies (red) as much. While high frequencies are still modified more than low frequencies, extensive alterations of low frequencies in the filter bank also affect brightness and contrast, thereby reducing image visibility (see Fig.\ref{fig:samples_filters}). Surprisingly, despite these modifications, the Card model proves to be the most robust against our attack, as seen in Tab.\ref{tab:res_all}. This indicates that our attack fails to find a combination capable of fooling the Card model, highlighting its resilience against various image corruptions compared to methods like Prime and Augmix. \\
\textbf{ImageNet:} Through the application of MUFIA, we find that altering low frequencies is crucial for inducing misclassification across all models on this dataset. This observation highlights that frequency bias is influenced not only by the training method but also by the image content itself. Consequently, our attack proves necessary as it can dynamically modify frequencies based on the specific dataset and on the evaluated model. In Figure \ref{fig:median_filter}, we observe a reduction in all models' low frequencies, although not as prominently as in the Card model \cite{card} on the CIFAR-10/100 dataset. This type of change typically affects brightness or causes blurring rather than the significant visibility reduction observed in the Card model. Similarly, Figure \ref{fig:samples_filters} demonstrates a degree of blur and brightness alteration across all models. Table \ref{tab:res_all} reveals that our attack algorithm affects all models on this dataset. We hypothesize that DeiT-B's slightly better performance may be attributed to its higher-capacity architecture \cite{vit_robust_1,vit_robust_2} compared to the rest of the models. Nevertheless, when subject to our attack, the accuracy of this model is significantly reduced by $90\%$ compared to its accuracy on the Common Corruption dataset. This highlights that relying solely on a larger capacity model is insufficient to effectively counter the novel corruptions introduced by our attack. Instead, these architectures may need to be complemented with techniques that mitigate over-fitting and promote generalization, such as the approach employed by Card \cite{card}.

\section{Conclusion}
In conclusion, while deep learning models have shown remarkable performance in various tasks, their effectiveness in real-world scenarios is limited by their inability to handle diverse image corruptions. Existing methods that address corruptions are often tailored to specific benchmark datasets, overlooking broader challenges. To simplify the analysis of model robustness, we propose MUFIA, an algorithm that identifies frequency combinations that the model is vulnerable to. Our findings indicate that data augmentation-based methods are susceptible to blur and brightness changes, while even state-of-the-art models struggle with low visibility-based corruptions. This highlights the need for comprehensive approaches based on bigger architectures like vision transformers \cite{imagenet_sota} or newer training methods like Card \cite{card} are needed to enhance model robustness against a wider range of real-world image corruptions.

\bibliography{egbib}

\begin{thebibliography}{34}
\providecommand{\natexlab}[1]{#1}
\providecommand{\url}[1]{\texttt{#1}}
\expandafter\ifx\csname urlstyle\endcsname\relax
  \providecommand{\doi}[1]{doi: #1}\else
  \providecommand{\doi}{doi: \begingroup \urlstyle{rm}\Url}\fi

\bibitem[* et~al.(2018)*, *, and Bethge]{decision_black_box}
Wieland~Brendel *, Jonas~Rauber *, and Matthias Bethge.
\newblock Decision-based adversarial attacks: Reliable attacks against
  black-box machine learning models.
\newblock \emph{ICLR}, 2018.

\bibitem[Ahmed et~al.(1974)Ahmed, Natarajan, and Rao]{dct_orig}
N.~Ahmed, T.~Natarajan, and K.R. Rao.
\newblock Discrete cosine transform.
\newblock \emph{IEEE Transactions on Computers}, 1974.

\bibitem[Benz et~al.(2021)Benz, Ham, Zhang, Karjauv, and Kweon]{vit_robust_1}
Philipp Benz, Soomin Ham, Chaoning Zhang, Adil Karjauv, and In~So Kweon.
\newblock Adversarial robustness comparison of vision transformer and mlp-mixer
  to cnns.
\newblock \emph{arXiv preprint arXiv:2110.02797}, 2021.

\bibitem[Bhattad et~al.(2020)Bhattad, Chong, Liang, Li, and
  Forsyth]{texture_attack}
Anand Bhattad, Min~Jin Chong, Kaizhao Liang, Bo~Li, and D.~A. Forsyth.
\newblock Unrestricted adversarial examples via semantic manipulation.
\newblock \emph{ICLR}, 2020.

\bibitem[Carlini and Wagner(2017)]{carlini_attack}
Nicholas Carlini and David Wagner.
\newblock Towards evaluating the robustness of neural networks.
\newblock \emph{IEEE Symposium on Security and Privacy}, 2017.

\bibitem[Croce et~al.(2020)Croce, Andriushchenko, Sehwag, Debenedetti,
  Flammarion, Chiang, Mittal, and Hein]{robustbench}
Francesco Croce, Maksym Andriushchenko, Vikash Sehwag, Edoardo Debenedetti,
  Nicolas Flammarion, Mung Chiang, Prateek Mittal, and Matthias Hein.
\newblock Robustbench: a standardized adversarial robustness benchmark.
\newblock \emph{arXiv preprint arXiv:2010.09670}, 2020.

\bibitem[Cubuk et~al.(2019)Cubuk, Zoph, Mané, Vasudevan, and Le]{auto_augment}
Ekin~D. Cubuk, Barret Zoph, Dandelion Mané, Vijay Vasudevan, and Quoc~V. Le.
\newblock Autoaugment: Learning augmentation strategies from data.
\newblock \emph{IEEE CVPR}, 2019.

\bibitem[Deininger et~al.(2022)Deininger, Stimpel, Yuce, Abbasi-Sureshjani,
  Sch{\"o}nenberger, Ocampo, Korski, and Gaire]{vit_robust_2}
Luca Deininger, Bernhard Stimpel, Anil Yuce, Samaneh Abbasi-Sureshjani, Simon
  Sch{\"o}nenberger, Paolo Ocampo, Konstanty Korski, and Fabien Gaire.
\newblock A comparative study between vision transformers and cnns in digital
  pathology.
\newblock \emph{arXiv preprint arXiv:2206.00389}, 2022.

\bibitem[Deng et~al.(2009)Deng, Dong, Socher, Li, Li, and Fei-Fei]{imagenet}
Jia Deng, Wei Dong, Richard Socher, Li-Jia Li, Kai Li, and Li~Fei-Fei.
\newblock Imagenet: A large-scale hierarchical image database.
\newblock \emph{IEEE CVPR}, 2009.

\bibitem[Diffenderfer et~al.(2021)Diffenderfer, Bartoldson, Chaganti, Zhang,
  and Kailkhura]{card}
James Diffenderfer, Brian Bartoldson, Shreya Chaganti, Jize Zhang, and Bhavya
  Kailkhura.
\newblock A winning hand: Compressing deep networks can improve
  out-of-distribution robustness.
\newblock \emph{NeurIPS}, 2021.

\bibitem[Geirhos et~al.(2019)Geirhos, Rubisch, Michaelis, Bethge, Wichmann, and
  Brendel]{shape_texture}
Robert Geirhos, Patricia Rubisch, Claudio Michaelis, Matthias Bethge, Felix~A
  Wichmann, and Wieland Brendel.
\newblock Imagenet-trained cnns are biased towards texture; increasing shape
  bias improves accuracy and robustness.
\newblock \emph{ICLR}, 2019.

\bibitem[He et~al.(2016)He, Zhang, Ren, and Sun]{resnet}
Kaiming He, Xiangyu Zhang, Shaoqing Ren, and Jian Sun.
\newblock Deep residual learning for image recognition.
\newblock \emph{IEEE CVPR}, 2016.

\bibitem[Hendrycks and Dietterich(2019)]{cc}
Dan Hendrycks and Thomas Dietterich.
\newblock Benchmarking neural network robustness to common corruptions and
  perturbations.
\newblock \emph{ICLR}, 2019.

\bibitem[Hendrycks et~al.(2020)Hendrycks, Mu, Cubuk, Zoph, Gilmer, and
  Lakshminarayanan]{Augmix}
Dan Hendrycks, Norman Mu, Ekin~D. Cubuk, Barret Zoph, Justin Gilmer, and Balaji
  Lakshminarayanan.
\newblock {AugMix}: A simple data processing method to improve robustness and
  uncertainty.
\newblock \emph{ICLR}, 2020.

\bibitem[Kireev et~al.(2022)Kireev, Andriushchenko, and
  Flammarion]{maksym_adv_cc}
Klim Kireev, Maksym Andriushchenko, and Nicolas Flammarion.
\newblock On the effectiveness of adversarial training against common
  corruptions.
\newblock \emph{UAI}, 2022.

\bibitem[Krizhevsky et~al.(2009)Krizhevsky, Hinton, et~al.]{cifar10/100}
Alex Krizhevsky, Geoffrey Hinton, et~al.
\newblock Learning multiple layers of features from tiny images.
\newblock 2009.

\bibitem[Li et~al.(2023)Li, Ortega~Caro, Rusak, Brendel, Bethge, Anselmi,
  Patel, Tolias, and Pitkow]{bethge_freq}
Zhe Li, Josue Ortega~Caro, Evgenia Rusak, Wieland Brendel, Matthias Bethge,
  Fabio Anselmi, Ankit~B Patel, Andreas~S Tolias, and Xaq Pitkow.
\newblock Robust deep learning object recognition models rely on low frequency
  information in natural images.
\newblock \emph{PLOS Computational Biology}, 2023.

\bibitem[Machiraju and Channappayya(2018)]{mine_av}
Harshitha Machiraju and Sumohana~S. Channappayya.
\newblock An evaluation metric for object detection algorithms in autonomous
  navigation systems and its application to a real-time alerting system.
\newblock \emph{IEEE ICIP}, 2018.

\bibitem[Machiraju et~al.(2022)Machiraju, Choung, Herzog, and
  Frossard]{mine_bio}
Harshitha Machiraju, Oh-Hyeon Choung, Michael~H Herzog, and Pascal Frossard.
\newblock Empirical advocacy of bio-inspired models for robust image
  recognition.
\newblock \emph{arXiv preprint arXiv:2205.09037}, 2022.

\bibitem[Madry et~al.(2018)Madry, Makelov, Schmidt, Tsipras, and
  Vladu]{madry_adv_train}
Aleksander Madry, Aleksandar Makelov, Ludwig Schmidt, Dimitris Tsipras, and
  Adrian Vladu.
\newblock Towards deep learning models resistant to adversarial attacks.
\newblock \emph{ICLR}, 2018.

\bibitem[Michaelis et~al.(2019)Michaelis, Mitzkus, Geirhos, Rusak, Bringmann,
  Ecker, Bethge, and Brendel]{bethge_av}
Claudio Michaelis, Benjamin Mitzkus, Robert Geirhos, Evgenia Rusak, Oliver
  Bringmann, Alexander~S Ecker, Matthias Bethge, and Wieland Brendel.
\newblock Benchmarking robustness in object detection: Autonomous driving when
  winter is coming.
\newblock \emph{arXiv preprint arXiv:1907.07484}, 2019.

\bibitem[Mintun et~al.(2021)Mintun, Kirillov, and Xie]{cc_better}
Eric Mintun, Alexander Kirillov, and Saining Xie.
\newblock On interaction between augmentations and corruptions in natural
  corruption robustness.
\newblock \emph{NeurIPS}, 2021.

\bibitem[Modas et~al.(2022)Modas, Rade, {Ortiz-Jim\'enez}, {Moosavi-Dezfooli},
  and Frossard]{Prime}
Apostolos Modas, Rahul Rade, Guillermo {Ortiz-Jim\'enez}, Seyed-Mohsen
  {Moosavi-Dezfooli}, and Pascal Frossard.
\newblock Prime: A few primitives can boost robustness to common corruptions.
\newblock \emph{ECCV}, 2022.

\bibitem[Oquab et~al.(2023)Oquab, Darcet, Moutakanni, Vo, Szafraniec, Khalidov,
  Fernandez, Haziza, Massa, El-Nouby, Howes, Huang, Xu, Sharma, Li, Galuba,
  Rabbat, Assran, Ballas, Synnaeve, Misra, Jegou, Mairal, Labatut, Joulin, and
  Bojanowski]{dino}
Maxime Oquab, Timothée Darcet, Theo Moutakanni, Huy~V. Vo, Marc Szafraniec,
  Vasil Khalidov, Pierre Fernandez, Daniel Haziza, Francisco Massa, Alaaeldin
  El-Nouby, Russell Howes, Po-Yao Huang, Hu~Xu, Vasu Sharma, Shang-Wen Li,
  Wojciech Galuba, Mike Rabbat, Mido Assran, Nicolas Ballas, Gabriel Synnaeve,
  Ishan Misra, Herve Jegou, Julien Mairal, Patrick Labatut, Armand Joulin, and
  Piotr Bojanowski.
\newblock Dinov2: Learning robust visual features without supervision, 2023.

\bibitem[O'Reilly et~al.(2022)O'Reilly, Bugler, Bhandari, Morrison, and
  Pardo]{voicebox}
Patrick O'Reilly, Andreas Bugler, Keshav Bhandari, Max Morrison, and Bryan
  Pardo.
\newblock Voiceblock: Privacy through real-time adversarial attacks with
  audio-to-audio models.
\newblock \emph{NeurIPS}, 2022.

\bibitem[Ortiz-Jimenez et~al.(2020)Ortiz-Jimenez, Modas, Moosavi, and
  Frossard]{guille}
Guillermo Ortiz-Jimenez, Apostolos Modas, Seyed-Mohsen Moosavi, and Pascal
  Frossard.
\newblock Hold me tight! influence of discriminative features on deep network
  boundaries.
\newblock \emph{NeurIPS}, 2020.

\bibitem[Ortiz-Jiménez et~al.(2021)Ortiz-Jiménez, Modas, Moosavi-Dezfooli,
  and Frossard]{guille_survey}
Guillermo Ortiz-Jiménez, Apostolos Modas, Seyed-Mohsen Moosavi-Dezfooli, and
  Pascal Frossard.
\newblock Optimism in the face of adversity: Understanding and improving deep
  learning through adversarial robustness.
\newblock \emph{Proceedings of the IEEE}, 2021.

\bibitem[Tian et~al.(2022)Tian, Wu, Dai, Hu, and Jiang]{imagenet_sota}
Rui Tian, Zuxuan Wu, Qi~Dai, Han Hu, and Yugang Jiang.
\newblock Deeper insights into vits robustness towards common corruptions.
\newblock \emph{arXiv preprint arXiv:2204.12143}, 2022.

\bibitem[Wallace(1992)]{jpeg_orig}
G.K. Wallace.
\newblock The jpeg still picture compression standard.
\newblock \emph{IEEE Transactions on Consumer Electronics}, 1992.

\bibitem[Wang et~al.(2022)Wang, Bochkovskiy, and Liao]{yolo_new}
Chien-Yao Wang, Alexey Bochkovskiy, and Hong-Yuan~Mark Liao.
\newblock Yolov7: Trainable bag-of-freebies sets new state-of-the-art for
  real-time object detectors.
\newblock \emph{arXiv preprint arXiv:2207.02696}, 2022.

\bibitem[Yin et~al.(2019)Yin, Gontijo~Lopes, Shlens, Cubuk, and
  Gilmer]{fourier_yin}
Dong Yin, Raphael Gontijo~Lopes, Jon Shlens, Ekin~Dogus Cubuk, and Justin
  Gilmer.
\newblock A fourier perspective on model robustness in computer vision.
\newblock \emph{NeurIPS}, 2019.

\bibitem[Zhang et~al.(2018)Zhang, Isola, Efros, Shechtman, and Wang]{lpips}
Richard Zhang, Phillip Isola, Alexei~A Efros, Eli Shechtman, and Oliver Wang.
\newblock The unreasonable effectiveness of deep features as a perceptual
  metric.
\newblock \emph{IEEE CVPR}, 2018.

\bibitem[Zhang et~al.(2021)Zhang, Zhao, Liu, Li, Cheng, Zheng, and
  Hu]{audio_loss_cite}
Weiyi Zhang, Shuning Zhao, Le~Liu, Jianmin Li, Xingliang Cheng, Thomas~Fang
  Zheng, and Xiaolin Hu.
\newblock Attack on practical speaker verification system using universal
  adversarial perturbations.
\newblock \emph{IEEE ICASSP}, 2021.

\bibitem[Zhao et~al.(2020)Zhao, Liu, and Larson]{ace_color}
Zhengyu Zhao, Zhuoran Liu, and Martha Larson.
\newblock Adversarial color enhancement: Generating unrestricted adversarial
  images by optimizing a color filter.
\newblock \emph{arXiv preprint arXiv:2002.01008}, 2020.

\end{thebibliography}

\clearpage
\appendix

\section{Ablation}
\label{sec:Ablation}
MUFIA has a few hyperparameters (Algo.\ref{algo:original}), which could change the performance of the algorithm. In this current Section, we conduct a hyperparameter study to analyze the effect of different settings. For all our analysis in the following Section, we use a standard trained ResNet20 \cite{resnet} model on the CIFAR-10 dataset. 

\subsection{Effect of $N_{iters}$}
\begin{figure}[htbp!]
    \centering
    \includegraphics[width=\textwidth]{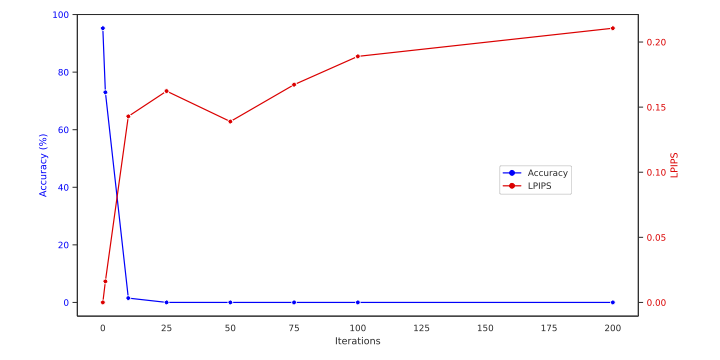}
    \caption{Accuracy and LPIPS vs. Iterations on CIFAR10.}
    \label{fig:iters_vary}
\end{figure}

Our attack algorithm, MUFIA, includes a hyper-parameter called $N_{iters}$, which determines the number of iterations the attack algorithm undergoes. 
We conduct a study to analyze the impact of this parameter on accuracy and LPIPS by varying the attack iterations across different values: from 0 to 200. The value of 0 represents no attack being applied. 
As shown in Fig. \ref{fig:iters_vary}, with just 1 iteration of the attack algorithm, we can reduce the accuracy by $10\%$. Further, $10$ iterations are sufficient to fool the model entirely. Additionally, the LPIPS values are well below the values seen in the common corruptions dataset \cite{cc} of $0.45$\cite{maksym_adv_cc}, showing that the adversarial images produced are still semantically similar to the original image.
 % \vspace{-2mm}

\subsection{Block size:}
\begin{figure}[h!]
    \centering
    \includegraphics[width=\textwidth]{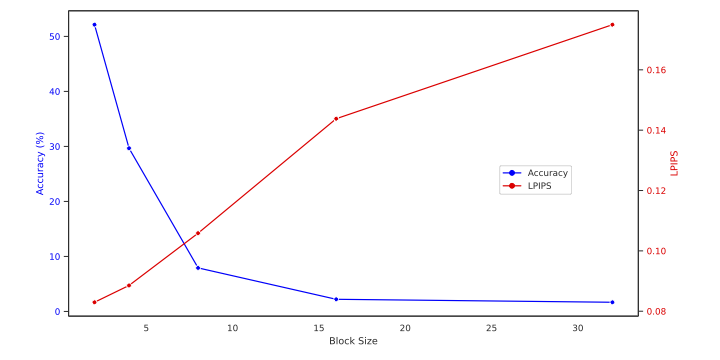}
    \caption{Accuracy and LPIPS vs. Block sizes on CIFAR10.}
    \label{fig:block_vary}
\end{figure}

Our attack algorithm MUFIA creates an adversarial filter bank in the DCT domain. The computation of the DCT transform increases as the size of the image blocks increases \cite{dct_orig}. While we set moderate block sizes for our experiments in Sec.\ref{sec:methodology}, we investigate the effect of different variations of the parameter choices. We experiment with $\lambda=10$ and $\kappa=0.99$ and run our attack for $10$ iterations. We then vary our block size in powers of 2, from 2 to 32. We plot the accuracy and LPIPS trade-off for different block sizes in Fig. \ref{fig:block_vary}. As the block size increases, we gain more degrees of freedom, making it easier to attack the model. Further, with larger block sizes to attack the model, there is more precise control over various frequency bands, hence 
more changes can be introduced in the image, corresponding to the increasing LPIPS value with increasing block size in Fig.\ref{fig:block_vary}. We see that even with a block size of $16$, which is half of image size in CIFAR10, our algorithm has sufficient degrees of freedom to find optimal filter banks that can cause misclassification.\\

% For completeness, we also show that on ImageNet, since it is computationally expensive we cannot use the whole image size 224 as our block size. However, even for lower block sizes as shown in Fig.\ref{fig:imagenet_blocks_big}, we do not see any drastic block artifacts in any of the images. This shows that algorithm produces, smooth changes in  

\subsection{Hyper-parameter Tradeoffs: $\kappa$ vs. $\lambda$ }
\label{sec:kappa_vs_lambda}
\begin{figure}[h!]
\centering
  \centering
  \includegraphics[width=\textwidth]{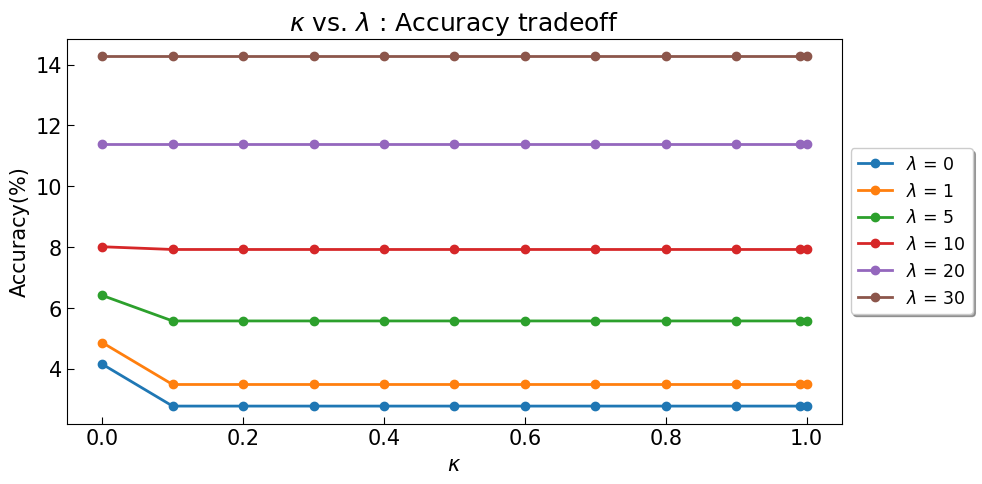}
  \captionof{figure}{$\kappa$ and $\lambda$ accuracy trade-offs. For a given value of $\lambda$, we plot the accuracy variations with changes in $\kappa$. We see with increasing $\lambda$, the accuracy of the model steadily increases, and beyond $\lambda=5$, changing $\kappa$ does not effect the accuracy of the model. This shows us that when such high importance is placed on the image similarity part of the loss, our attack MUFIA is unable to produce modifications in the image that can actually fool the model. 
  For $\lambda \leq 5$, we see that increasing $\kappa$, forces MUFIA to further fool the model. This is in line with the purpose of the thresholding parameter $\kappa$, that we explained in Sec.\ref{sec:methodology}. }
  \label{fig:tradeoff_accs}
\end{figure}

\begin{figure}[h!]
  \centering
  \includegraphics[width=\textwidth]{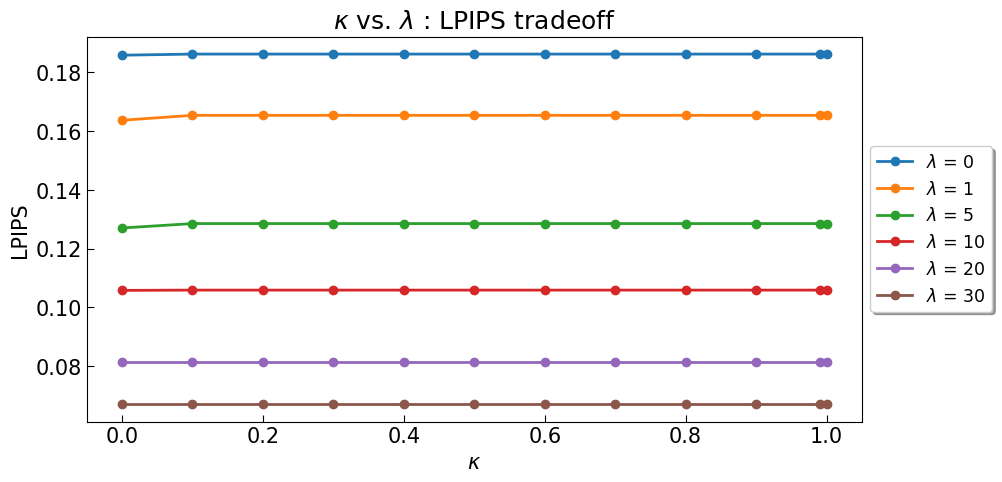}
  \captionof{figure}{$\kappa$ and $\lambda$ LPIPS trade-offs. For a given value of $\lambda$, we plot the accuracy variations with changes in $\kappa$. We see with increasing $\lambda$, the LPIPS of the model steadily decreases, and beyond $\lambda=5$, changing $\kappa$ does not effect the LPIPS of the model. This shows us that when such high importance is placed on the image similarity part of the loss, our attack MUFIA indeed produces adversarial images that are very similar to the original image. However, as we have seen in Fig.\ref{fig:tradeoff_accs}, the images produced at these higher values of $\lambda$, have lower rates of success in fooling the model.}
  \label{fig:tradeoff_lpips}
\end{figure}

Ideally, we expect that with a higher value of $\kappa$, we get a higher fooling rate. However, the paramter $\lambda$ acts as a trade-off parameter to maintain image similarity. To analyze this trade-off, we vary our $\lambda$ and $\kappa$ values. To reduce computational needs, we use just $10$ iterations of our attack and a block size of 8. As shown in Fig. \ref{fig:tradeoff_accs}, for a fixed value of $\kappa$, with increasing $\lambda$, we observe that the accuracy steadily increases while the LPIPS value decreases as shown in Fig. \ref{fig:tradeoff_lpips}. As expected, this behavior coincides with a trade-off between accuracy and image similarity. Additionally, in Fig. \ref{fig:tradeoff_accs}, we observe that for any given value of $\lambda$, increasing the value of $\kappa$ does not make any further difference. This indicates that it is impossible to cause a further decrease in accuracy within such a strict image similarity constraint.
Furthermore, for $\lambda=0$, $\kappa=0.2$ is sufficient to cause misclassification for this model. We could safely say that there must be sufficiently small values of $\kappa$ and $\lambda$ that ensure almost imperceptible perturbations. For our experiments, we used much larger values to ensure that the models are definitely fooled. However, with sufficient hyperparameter search, more optimal values can be found, ensuring very low LPIPS values.

\clearpage

\section{Filter banks}
The entirety of the DCT spectrum is quite hard to interpret. It is difficult to say what each frequency component represents in a natural image. However, we can group frequencies easily and say that changing low frequency would affect brightness information or high frequency would affect the edges, respectively. As, looking at each filter bank for each image would be quite complex, so we first look at more straightforward experiments to demonstrate the efficacy of our algorithm, MUFIA. The goal of our experiment to check if our algorithm's can even produces changes in relevant frequency bands or does it just induce random changes. 

For our experiments, we choose a $2 \times 2$ DCT filter bank with the following representation: 
\[
\mathbf{Q} = \begin{bmatrix}
q_{0,0} & q_{0,1} \\
q_{1,0} & q_{1,1} \\
\end{bmatrix}
\]
Each of the elements represents the following : 
\begin{itemize}
    \item $q_{0,0}$ controls the average brightness of all the pixels in the image.
    \item $q_{0,1}$ controls the vertical change in pixels.
    \item $q_{1,0}$ controls the horizontal change in pixels. 
    \item $q_{1,1}$ controls the diagonal change in pixels.
\end{itemize}

% , pretrained on the CIFAR10 dataset. 
% \red{decision-black box attack...ie just flip label so c=maxf(x)}
We use our attack MUFIA, to predict the filter bank $D$, such that it can flip the decision of the model. To recap our adversarial loss was defined as follows:
$\mathcal{L}_{adv}$ loss : 
\begin{gather}
    \min_{Q} \mathcal{L}_{adv} =  \min_{Q} ({  \max(\mathcal{S}_{cos}(f(\hat{x}), c) + \kappa, 0) }) 
\end{gather}
Originally, in the above equation, $c$ is the defined one hot encoding  ground truth label ($t$). It can represented as :
% \centering
\begin{center}
$c = \mqty[c_{1} & c_{2} & \cdots & c_{n}]$ 
\hspace{2pt} where, 
$c_i :=
\begin{cases}
1 &\text{if } i = t, \\
0 &\text{else }
\end{cases}
$
\end{center}

For the purposes of this experiment, we redefine $c$ as a one hot encoding of the model's original predicted label on the image $x$, i.e.,
\begin{center}
    $c_i :=
\begin{cases}
1 &\text{if } i = \argmax f(x), \\
0 &\text{else }
\end{cases}
$
\end{center}
    
By doing so, the goal is to find $\hat{x}$, such that the label predicted on this new image is different from the label predicted on the original image $x$. This is done, while keeping $\hat{x}$ visually similar to the original image $x$.
\footnote{This is similar to decision based black box attack \cite{decision_black_box}.}

For our experiments in the following subsections, we use a standard ResNet-50 with 10 class outputs and set $\lambda=0$ to solely observe our adversarial algorithm's effects. We also set $\kappa=0.99$ for our $\mathcal{L}_{adv}$ loss function. Note that our DCT filter bank $Q$ is applied in a block-wise fashion on blocks of size $2 \times 2$ from the image.

% The pre-training or nature of the network does not make any difference to our results. 

% For all our experiments, we set the true label to be the prediction made by the network in the unchanged image. In essence, we force MUFIA to create an adversarial image such that the network does not have the same label as its original prediction. In this sense, the 'true' label for random pattern images is no longer necessary; hence, the training on the network does not make any difference to our results. We only want to look at the efficacy of our algorithm and not the inherent bias of any model.

% \clearpage
\subsection{$q_{0,0}$ element}
\begin{figure}[htbp!]
    \centering
    \includegraphics[width=\textwidth]{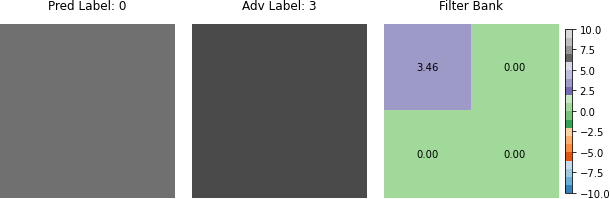}
    \caption{Original image on the left, generated using a single color and no variation in pixels. The image in the middle shows the adversarial image, with lower brightness produced by the application of MUFIA. On the right, we see the corresponding filter bank $Q$ produced by MUFIA for the given original image.}
    \label{fig:d00}
\end{figure}

\begin{figure}[htbp!]
    \centering
    \includegraphics[scale=0.5]{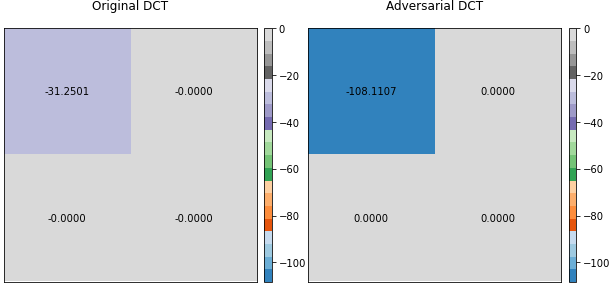}
    \caption{$2 \times 2$ image block DCT elements for images shown in Fig.\ref{fig:d00}. The Filter bank is applied element-wise to the original DCT elements, which results in the adversarial DCT elements seen on the right. As shown in the original DCT on the left, the brightness was low, to begin with. However, with the application of our filter bank, it is further suppressed, creating an even darker image as seen in Fig.\ref{fig:d00}. }
    \label{fig:d00_dct}
\end{figure}

We use an input image of a single color with no pixel variations. The input image we introduce to the model is shown in Fig.\ref{fig:d00}. The network predicts the label of such an image as $0$. In order to change the label predicted by the image, our algorithm MUFIA needs to be effective in changing the $q_{0,0}$ element of the filter bank. 
As shown in Fig.\ref{fig:d00}, upon applying MUFIA, the $q_{0,0}$ component is indeed altered. As shown in Fig.\ref{fig:d00_dct}, Since the $\{0, 0\}$ component of the original image DCT, is already low to begin with, the application of our filter bank results in even lower DC component, resulting in lower average brightness of the image. This is indeed resultant effect in the adversarial image seen in Fig.\ref{fig:d00}. 
\footnote{Note for all our experiments due to the redundancy in information across all the blocks for the generated images we only plot one $2 \times 2$ block of the DCT of the original and final images for easier visualization. }$^{,\thinspace}$\footnote{The small negative sign next to some of the zeros in the DCTs is due to small numerical errors in Python. Realistically, they should be zero but numerically they are in the order of $1e-10$.}

% we see that the image's overall brightness is reduced. We also see that the generated filter bank promotes this reduction in brightness.

% \newpage
\subsection{$q_{0,1}$ element}
\begin{figure}[htbp!]
    \centering
    \includegraphics[width=\textwidth]{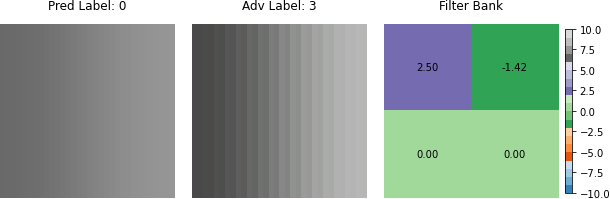}
    \caption{Original image on the left generated by creating variation along the spatial x-axis of the image. The image in the middle shows the adversarial image produced by the application of MUFIA, with lower average brightness and more pronounced vertical variations. On the right, we see the corresponding filter bank $Q$ produced by MUFIA for the original image.}
    \label{fig:d01}
\end{figure}

\begin{figure}[htbp!]
    \centering
    \includegraphics[scale=0.5]{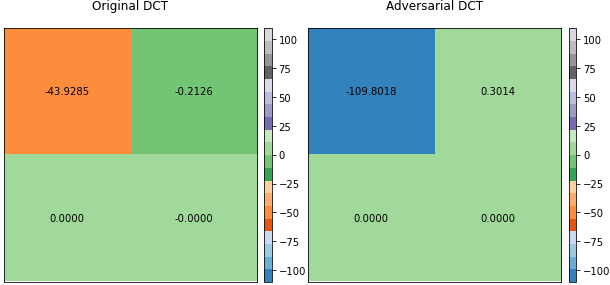}
    \caption{$2 \times 2$ image block DCT elements for images shown in Fig.\ref{fig:d01}. The Filter bank is applied element-wise to the original DCT elements, which results in the net adversarial DCT elements seen on the right. As we can see from the adversarial DCT elements, the final image should, on average, have lower brightness and more pronounced variation along the spatial x-axis. This is in accordance with the figure we find in Fig.\ref{fig:d01}.}
    \label{fig:d01_dct}
\end{figure}

We create an input image with only vertical variation in pixels as shown in Fig.\ref{fig:d01}. The network predicts the label of such an image as $0$. In order to change the label predicted by the image, our algorithm MUFIA needs to be effective in changing the $q_{0,1}$ element of the filter bank. This should cause change in the how sharp/blurred the vertical line variation becomes. In addition, MUFIA could also modify the $q_{0,0}$ element and cause overall changes in brightness across the image. 
As shown in Fig.\ref{fig:d01}, upon applying MUFIA, the $q_{0,1}$ component is indeed altered, along with the $q_{0,0}$ component. As shown in Fig.\ref{fig:d00_dct}, Since the $\{0, 0\}$ component of the original image DCT, is already low to begin with, the application of our filter bank results in even lower DC component, resulting in lower average brightness of the image. In addition, since the $\{0,1\}$ element of the original DCT is low to begin with, applying our predicted filter bank $Q$, results in a more positive and higher component. This would cause a more pronounced effect in vertical variations, similar to resultant adversarial image seen in Fig.\ref{fig:d01_dct}.

% \clearpage
\subsection{$q_{1,0}$ element}

We create an input image with only horizontal variation in pixels as shown in Fig.\ref{fig:d01}. The network predicts the label of such an image as $0$. In order to change the label predicted by the image, our algorithm MUFIA needs to be effective in changing the $q_{1,0}$ element of the filter bank. This should cause change in the how sharp/blurred the horizontal line variation becomes. In addition, MUFIA could also modify the $q_{0,0}$ element and cause overall changes in brightness across the image. 
As shown in Fig.\ref{fig:d01}, upon applying MUFIA, the $q_{1,0}$ component is indeed altered, along with the $q_{0,0}$ component. As shown in Fig.\ref{fig:d00_dct}, Since the $\{0, 0\}$ component of the original image DCT, is already low to begin with, the application of our filter bank results in even lower DC component, resulting in lower average brightness of the image. In addition, since the $\{1,0\}$ element of the original DCT is low to begin with, applying our predicted filter bank $Q$, results in a more negative and but higher magnitude component. This would cause a more pronounced effect in horizontal variations, similar to resultant adversarial image seen in Fig.\ref{fig:d10_dct}. 

\begin{figure}[h!]
    \centering
    \includegraphics[width=\textwidth]{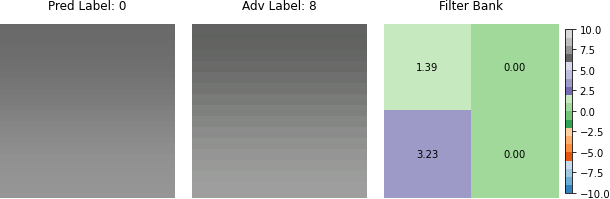}
    \caption{Original image on the left generated by creating variation along the spatial y-axis of the image. The image in the middle shows the adversarial image produced by the application of MUFIA. On the right, we see the corresponding filter bank $Q$ produced by MUFIA for the original image.}
    \label{fig:d10}
\end{figure}

\begin{figure}[h!]
    \centering
    \includegraphics[scale=0.5]{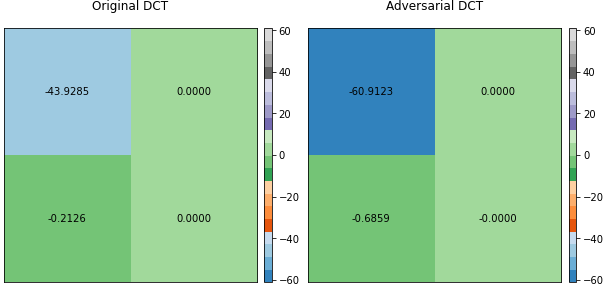}
    \caption{$2 \times 2$ image block DCT elements for images shown in Fig.\ref{fig:d10}. The Filter bank is applied element-wise to the original DCT elements, which results in the net adversarial DCT elements seen on the right. As we can see from the adversarial DCT elements, the final image should, on average, have lower brightness and more pronounced variation along the spatial y-axis. This is in accordance with the figure we find in Fig.\ref{fig:d10}.}
    \label{fig:d10_dct}
\end{figure}

% \clearpage
\subsection{$q_{1,1}$ element}

\begin{figure}[htbp!]
    \centering
    \includegraphics[width=\textwidth]{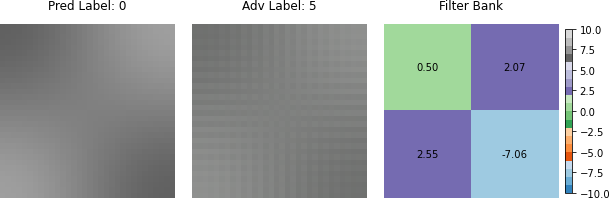}
    \caption{Original image on the left generated by creating variation along the all axes. The image in the middle shows the adversarial image produced by the application of MUFIA, with more pronounced effects across the horizontal, diagonal and vertical directions. On the right, we see the corresponding filter bank $Q$ produced by MUFIA for the original image.}
    \label{fig:d11}
\end{figure}

\begin{figure}[htbp!]
    \centering
    \includegraphics[scale=0.5]{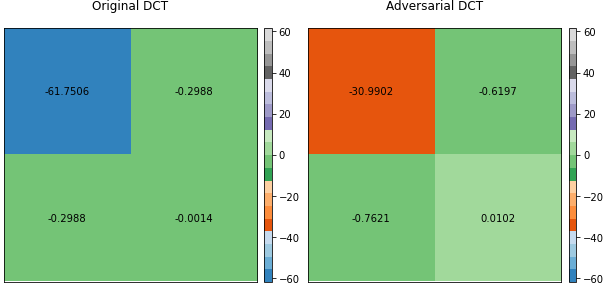}
    \caption{$2 \times 2$ image block DCT elements for images shown in Fig.\ref{fig:d11}. The Filter bank is applied element-wise to the original DCT elements, which results in the net adversarial DCT elements seen on the right. As we can see from the adversarial DCT elements, the final image should, on average, have relatively higher brightness and have more pronounced variation along both axes and the diagonals. This is in accordance with the figure we find in Fig.\ref{fig:d11}.}
    \label{fig:d11_dct}
\end{figure}

Finally, we put our algorithm to test by introducing both horizontal and vertical changes in the input changes as shown in Fig.\ref{fig:d11}.
The network predicts the label of such an image as $0$. In order to change the label predicted by the image, our algorithm MUFIA find a suitable filter bank for all frequency components. 
As shown in Fig.\ref{fig:d11}, upon applying MUFIA, indeed all components of the filter bank are utilized by our algorithm. As shown in Fig.\ref{fig:d11_dct}, since the $\{0, 0\}$ component of the original image DCT, is already low to begin with, the application of our filter bank results in slightly lower magnitude of the DC component, resulting in slightly higher average brightness of the image. 
In addition, since the $\{1,0\}$ element of the original DCT is low to begin with, applying our predicted filter bank $Q$, results in a more negative and but higher magnitude component. This would cause a more pronounced effect in horizontal variations, similar to resultant adversarial image seen in Fig.\ref{fig:d11_dct}. A similar case, occurs for the vertical variations in the $\{0,1\}$ component.
For the diagonal component $\{1,1\}$ increases in magnitude in the resultant adversarial DCT, hence resulting in more pronounced effects across the diagonal in the final image as seen in Fig.\ref{fig:d11_dct}.

\subsection{Datasets}
From the above subsections, we can clearly see how MUFIA uses relevant frequency components to induce visible changes in the images. The amplification or attenuation seen in filter bank clearly corresponds to the changes in different horizontal/vertical or diagonal directions seen in the DCT matrices and the final adversarial images. This clearly, shows us that looking at these filter banks gives us valuable information about how much frequency component of the DCT matrix is modified in order to produce the adversarial image.

\begin{figure}[htbp!]
    \centering
    \includegraphics[width=\textwidth]{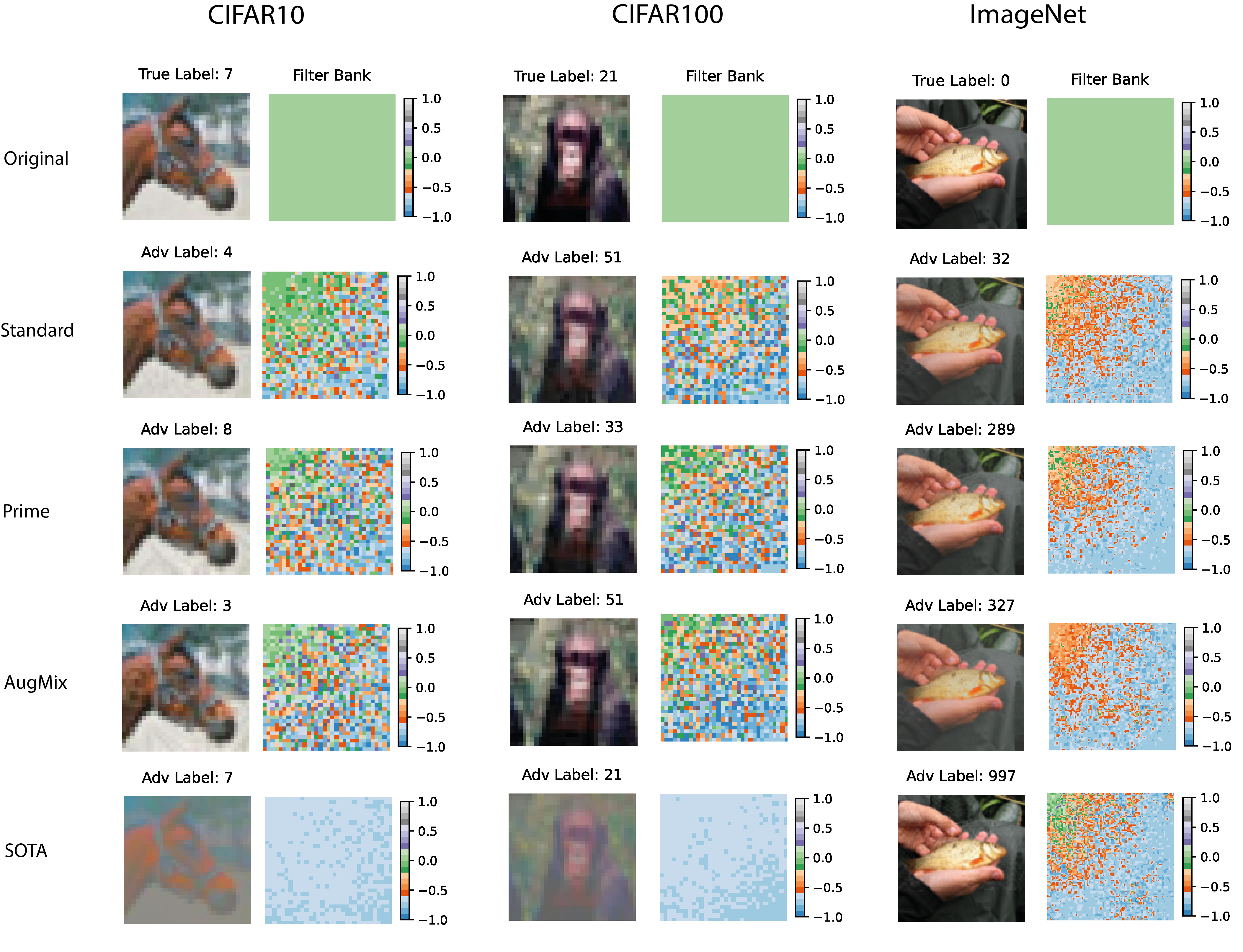}
    \caption{Images generated by our attack algorithm, MUFIA, for different datasets and models. Each column represents a dataset, and the first row shows the original, unchanged images with their true labels and what a filter bank that produces no change would look like. The following rows show the adversarial images generated by MUFIA for different models, with the predicted label at the top of each adversarial image. Note, for the CIFAR-10/100 columns, the SOTA model corresponds to Card \cite{card}, and for the ImageNet column, it corresponds to the DeiT-B model \cite{imagenet_sota}. To observe the subtle differences in the images themselves, we encourage the readers to zoom in. 
    To facilitate visual comparison, we apply a transformation using the $tanh(z-1)$ function to map the filter banks within the range of $(-1,1)$. In this transformed representation, a value of $0$ (green in the heatmap) indicates an unchanged frequency component $(z=1)$, positive values indicate amplification $(z>1)$ (purple, grey in the heatmap), and negative values indicate attenuation $(z<1)$ (orange/red, blues in the heatmap).}
    \label{fig:model_filters}
\end{figure}

While in the case of this toy example, we use simple $2 \times 2$ filter banks that are easy to interpret, in larger datasets it is very difficult to do this. However, for completeness, in Fig.\ref{fig:model_filters} we plot the filter banks obtained by our algorithm on images from different dataset and for different models (Sec.\ref{sec:experiments}). \footnote{Since, the DCT matrices are quite large and there is one per image block, it makes the visualization quite complex. Hence, we only stick to plotting the filter banks for these images.} We make the following observations from the filter banks:

% \red{the dct unlike the above examples is harder to look at for the whole 32x32 or 224x224 is hard to interpret}

\subsubsection{CIFAR10/100}
Similar to our observations from the median filter banks in Sec. \ref{sec:experiments}, we find that, in general, for the CIFAR10 and 100 datasets on the SOTA model (Card \cite{card}), we see lower visibility of the object in the adversarial image. This effect is induced through suppression of low and high-frequency information as seen in the corresponding filter banks for the images.
% \red{this is induced through reducing all components in the frequency bank} 
For the other models, on these datasets, we see that they keep some of the lower frequency information unchanged and mostly vary the mid and higher frequency information. Specifically on the CIFAR10 dataset, we see that for both the PRIME \cite{Prime} and Augmix \cite{Augmix} models, there are changes along the mouth of the horse. These diagonal changes along the image are also seen in the corresponding filter bank as purple components along the diagonal. With the amplification of the diagonal elements (purple components), the shape of the horse of the mouth is highlighted further by our algorithm MUFIA.

\subsubsection{ImageNet}
On the ImageNet dataset, we notice some low-frequency components suppressed, resulting in lower brightness. Addiotionally the high-frequency information is suppressed further resulting in blurs in the final image. Specifically on the Standard, Prime, and Augmix models, we can clearly see changes in brightness and slight blurring, with corresponding changes to both high and low frequencies in the filter bank. For the SOTA model (DeiT-B \cite{imagenet_sota}), we see a grainy appearance due to changes in high-frequency information. However, unlike the other images, since there is no change in very low-frequency information, we do not see any changes in brightness.

\newpage
\clearpage
\section{Loss: Cosine vs. Cross Entropy}

% 1. Bad quality of generated images
% 2. This is because of CE doesnt care abt perception etc... as written below..
% 3. logits for cosine vs ce r different i.e 
% Cross entropy is extremely sensitive to changes in logits. small changes in logits cause drastic change in the loss,
% to cause 
% while cosine is way more stable
% % Logits for Cosine cause confusion not happening just pushimng away from boundary 

\begin{figure}[h]
    \centering
    \includegraphics[width=0.8\textwidth]{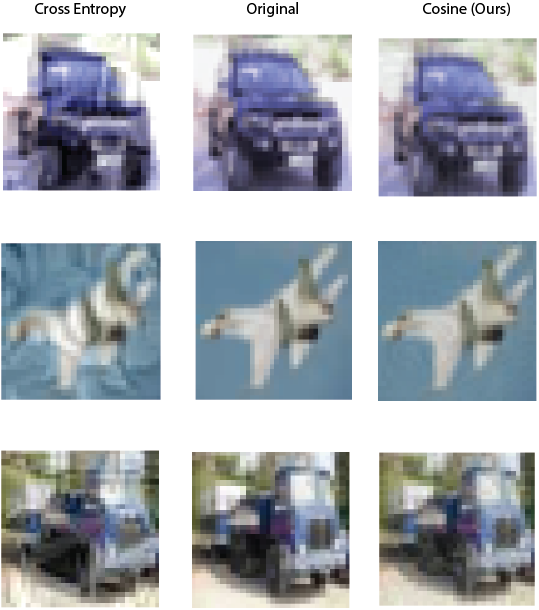}
    \caption{Images produced from the CIFAR10 dataset, modified by our attack algorithm using Cross entropy vs. Cosine loss for finding the adversarial filter bank. In the center column we show the original images present in the CIFAR10 dataset. In the rightmost column (Cosine), we represent images produced by our attack algorithm with the cosine-based loss term as mentioned in the main paper in Sec.\ref{sec:methodology}. For these images we set $\kappa=0.99$ and $\lambda=20$. In the leftmost column, the images are produced by replacing the cosine loss term in Sec.\ref{sec:methodology} with a traditional cross entropy-based loss used in adversarial attacks. We set $\lambda=20$ for these images also. 
    For all variations of the attack, it was performed on the CIFAR10 dataset with the standard trained ResNet-50 model.
    }
    \label{fig:loss_comp}
\end{figure}

Typically, adversarial attacks tend to use cross-entropy loss as their adversarial loss. However, as shown in Sec. \ref{sec:methodology}, we defined $\mathcal{L}_{adv}$ using cosine similarity, we may instead use the cross entropy loss. To test the effect of using cross entropy loss, we performed our attack MUFIA, with cross-entropy loss while keeping the same $L_{sim}$ loss for image similarity. We perform our experiment on CIFAR-10 with the ResNet-50 model and set $\lambda=20$. 
As shown in Fig. \ref{fig:loss_comp}, we find that the images generated using the cross-entropy loss bear low semantic similarity to the original image. Quite often the algorithm, destroys the image itself making it hard even for humans to classify them. Additionally, the accuracy of the model after this attack was also around $20\%$, which is quite high when compared to the complete misclassification ($0\%$) caused by our attack, when used with the cosine-based loss. We also find that for the cross-entropy loss based MUFIA, $\lambda$ had to be increased to $1000$ to produce more semantically similar images, but due to the trade-off nature of the $\lambda$ parameter, the accuracy of the model was also became very high ($75\%$). \\
This behaviour of our attack with cross entropy loss could be explained by its exponential nature.
Carlini et al.\cite{carlini_attack} have previously shown that due to the sensitive nature of cross-entropy loss, it oscillates between sub-optimal adversarial perturbations for a given input. With its exponential nature, it behaves in an overly greedy manner when gradient descent is performed. This causes the images produced to have too drastic changes like the ones we find in Fig. \ref{fig:loss_comp}. \\
In contrast, our loss $L_{adv}$ (Eq.\ref{eq:loss_adv}) with the cosine based term $S_{cos}$, turns to zero as soon as $S_{cos}$ becomes $-\kappa$. The loss stops updating hence preventing overly greedy solutions. In addition, the $S_{cos}$ term has a very stable gradient, which does not change drastically unlike the exponentially varying gradient of the cross entropy loss.\\
Hence, due to the stability of our loss in comparison to the cross entropy loss, we find that the former produces better quality images, while also being adversarial in nature. 
% In addition, to minimize the loss, the ideal solution is when $f(x) = -\alpha c$, where $\alpha \geq 0$. i.e. $f(x)$  and $c$ need to be opposite of each other. Combined with $L_{sim}$, we can find a trade-off solution such that there is some misalignment while remaining perceptually similar.
% \clearpage

\end{document}